# Compiling Possibilistic Networks : Alternative Approaches to Possibilistic Inference

**Raouia Ayachi**
LARODEC Laboratory
ISG, University of Tunis
Tunisia, 2000
raouia.ayachi@gmail.com

**Nahla Ben Amor**
LARODEC Laboratory
ISG, University of Tunis
Tunisia, 2000
nahla.benamor@gmx.fr

**Salem Benferhat**
CRIL-CNRS
University of Artois
France, 62307
benferhat@cril.univ-artois.fr

**Rolf Haenni**
RISIS
Bern University
Switzerland, CH-2501
rolf.haenni@bfh.ch

## Abstract

Qualitative possibilistic networks, also known as min-based possibilistic networks, are important tools for handling uncertain information in the possibility theory framework. Despite their importance, only the junction tree adaptation has been proposed for exact reasoning with such networks. This paper explores alternative algorithms using compilation techniques. We first propose possibilistic adaptations of standard compilation-based probabilistic methods. Then, we develop a new, purely possibilistic, method based on the transformation of the initial network into a possibilistic base. A comparative study shows that this latter performs better than the possibilistic adaptations of probabilistic methods. This result is also confirmed by experimental results.

## 1 INTRODUCTION

In possibility theory there are two different ways to define the counterpart of Bayesian networks. This is due to the existence of two definitions of possibilistic conditioning: product-based and min-based conditioning (Dubois and Prade, 1988). When we use the product form of conditioning, we get a possibilistic network close to the probabilistic one sharing the same features and having the same theoretical and practical results. However, this is not the case with min-based networks. In this paper, we are interested in the inference problem in multiply connected networks, which is known as a hard problem (Cooper, 1990). More precisely, we propose three compilation methods for min-based possibilistic networks.

The compilation of Bayesian networks is always considered as an important area. Recently, researchers have been interested in various kinds of exact and approximate Bayesian networks inference algorithms using compilation techniques (Darwiche, 2003) (Chavira and Darwiche, 2005) (Wachter and Haenni, 2007), etc.

Despite the importance of possibility theory, there is no compilation that has been proposed for possibilistic networks. This paper analyzes this issue by first adapting well-known compilation-based probabilistic inference approaches, namely the arithmetic circuit method (Darwiche, 2003) and the logical compilation of Bayesian Networks (Wachter and Haenni, 2007). Both of them are based on a network's encoding into a logical representation and a compilation into a target compilation language, namely Π-DNNF. From there, all possible queries are answered in polynomial time. The third method exploits results obtained on one hand in (Benferhat et al., 2002) that transforms a min-based possibilistic network into a possibilistic knowledge base, and on the other hand results obtained regarding compilation of possibilistic bases (Benferhat et al., 2007) in order to assure inference in polytime. This method that is purely possibilistic is flexible since it permits to exploit efficiently all the existing propositional compilers.

The rest of this paper is organized as follows: Section 2 gives a briefly background on possibility theory, possibilistic logic, possibilistic networks and introduces some compilation concepts. Section 3 is dedicated to possibilistic adaptations of compilation-based probabilistic inference methods. Section 4 presents a new inference method in possibilistic networks using compiled possibilistic knowledge bases. Experimental study is presented in Section 5.

## 2 BASIC CONCEPTS

### 2.1 POSSIBILITY THEORY

This subsection briefly recalls some elements of possibility theory, for more details we refer to (Dubois and Prade, 1988). Let $V = \{X_1, X_2, ..., X_N\}$ be a set of variables. We denote by $D_{X_i} = \{x_1, .., x_n\}$ the domain associated with the variable $X_i$. By $x_i$ we denote any instance of $X_i$. $\Omega$ denotes the universe of discourse, which is the Cartesian product of all variable domains in $V$. Each element $\omega \in \Omega$ is called a state of $\Omega$. The notion of possibility distribution denoted by $\pi$ is a mapping from the universe of discourse to the unit interval $[0, 1]$. To this scale, two interpretations can be attributed, a quantitative one when values have a real sense and a qualitative one when values reflect only an order between the different states of the world. This paper focuses on the qualitative interpretation of possibility theory.

Given a possibility distribution $\pi$, we can define a mapping grading the possibility measure of an event $\phi \subseteq \Omega$ by $\Pi(\phi) = max_{\omega \in \phi} \pi(\omega)$. $\Pi$ has a dual measure which is the necessity measure $N(\phi) = 1 - \Pi(\neg\phi)$.

Conditioning consists in modifying our initial knowledge, encoded by a possibility distribution $\pi$, by the arrival of a new certain piece of information $\phi \subseteq \Omega$. The qualitative interpretation of the scale $[0, 1]$ leads to the well known definition of min-conditioning (Hisdal, 1978), (Dubois and Prade, 1988):

$$\Pi(\psi \mid \phi) = \begin{cases} \Pi(\psi \wedge \phi) & if \ \Pi(\psi \wedge \phi) < \Pi(\phi) \\ 1 & \text{otherwise} \end{cases} \quad (1)$$

## 2.2 POSSIBILISTIC LOGIC

Possibilistic logic (Dubois et al., 1994) handles qualitative uncertainty in a logical setting. A possibilistic logic formula is a pair $(p, a)$ where $p$ is a *propositional* formula and $a$ its uncertainty degree which estimates to what extent it is certain that $p$ is true. The higher is the weight, the more certain is the formula. A possibilistic knowledge base $\Sigma$ is made up of a finite set of weighted formulas, i.e.,

$$\Sigma = \{(p_i, a_i), i = 1, .., n\} \quad (2)$$

where $a_i$ is the lower bound on $N(p_i)$.

Each possibilistic knowledge base induces a unique possibility distribution such that $\forall \ \omega \in \Omega$ and $\forall \ (p_i, a_i) \in \Sigma$:

$$\pi_\Sigma(\omega) = \begin{cases} 1 & if \ \omega \models p_i \\ 1 - max \{a_i : \omega \not\models p_i\} & \text{otherwise} \end{cases} \quad (3)$$

where $\models$ is propositional logic entailment.

## 2.3 POSSIBILISTIC NETWORKS

A min-based possibilistic network over a set of variables $V$, denoted by $\Pi G_{min}$ is composed of:
- a *graphical component* that is a DAG (Directed Acyclic Graph) where nodes represent variables and edges encode the links between the variables. The parent set of a node $X_i$ is denoted by $U_i = \{U_{i1}, U_{i2}, ..., U_{im}\}$. For any $u_i$ of $U_i$ we have $u_i = \{u_{i1}, u_{i2}, ..., u_{im}\}$ where $m$ is the number of parents of $X_i$. In what follows, we use $x_i, u_i, u_{ij}$ to denote, respectively, possible instances of $X_i, U_i$ and $U_{ij}$.
- a *numerical component* that quantifies different links. For every root node $X_i$ ($U_i = \varnothing$), uncertainty is represented by the a priori possibility degree $\Pi(x_i)$ of each instance $x_i \in D_{X_i}$, such that $max_{x_i}\Pi(x_i) = 1$. For the rest of the nodes ($U_i \neq \varnothing$) uncertainty is represented by the conditional possibility degree $\Pi(x_i|u_i)$ of each instances $x_i \in D_{X_i}$ and $u_i \in D_{U_i}$. These conditional distributions satisfy the following normalization condition: $max_{x_i}\Pi(x_i|u_i) = 1$, for any $u_i$.

The set of a priori and conditional possibility degrees in a min-based possibilistic network induce a unique joint possibility distribution defined by the following chain rule:

$$\pi_{min}(X_1, .., X_N) = \min_{i=1..N} \ \Pi(X_i \mid U_i) \quad (4)$$

## 2.4 COMPILATION CONCEPTS

A target compilation language is a class of formulas which is tractable for a set of *transformations* and *queries*. Compilation languages are compared in terms of their spatial efficiency via the *succinctness* criteria and also in terms of the set of *logical queries* and *transformations* they support in polynomial time (see (Darwiche and Marquis, 2002) for more details).

Within the most effective target compilation languages, we cite *the Decomposable Negation Normal Form (DNNF)* (Darwiche, 2001). This language is universal and presents a number of properties (*determinism*, *smoothness*, etc.) that makes it of a great interest. It supports a rich set of polynomial-time logical operations. To define DNNF, the starting point is *Negation Normal Form (NNF)* which is a set of propositional formulas where possible connectives are conjunctions, disjunctions and negations. A set of important properties may be imposed to NNF, such that:
- *Decomposability*: the conjuncts of any conjunction in NNF do not share variables.
- *Determinism*: two disjuncts of any disjunction in NNF are logically contradictory.
- *Smoothness*: the disjunct of any disjunction in NNF mentions the same variables.

These properties lead to a number of interesting subsets of NNF. Within these subsets, the language DNNF (Darwiche, 2001) is one of the most effective target compilation languages that supports the decomposability. We can also mention, the d-DNNF sat-isfying determinism, sd-DNNF satisfying smoothness and determinism, etc. Each compilation language supports some *queries* and *transformations* in polynomial time. In what follows we are in particular interested by *conditioning* and *forgetting* transformations (Darwiche and Marquis, 2002).

# 3 POSSIBILISTIC ADAPTATIONS OF COMPILATION-BASED PROBABILISTIC INFERENCE METHODS

There are several compilation methods which handle the inference problem in probabilistic graphical models. In this section, we first propose an adaptation of the arithmetic circuit method of (Darwiche, 2003). Then we will study one of its variants proposed in (Wachter and Haenni, 2007), namely the logical compilation of Bayesian Networks.

DNNF has been introduced for propositional language. Recall that in qualitative possibility theory, we basically manipulate two main operators Max and Min. These operators fully make sense when we deal with qualitative plausibility ordering. Therefore, we propose to define concepts of Π-DNNF (resp. Π-d-DNNF, Π-sd-DNNF) as adaptations of the DNNF language (resp. d-DNNF, sd-DNNF) (Darwiche, 2001) in the possibilistic setting (definition 1).

**Definition 1.** *A sentence in Π-DNNF is a rooted DAG where each leaf node is labeled with true, false or variable's instances and each internal node is labeled with max or min operators and can have arbitrarily several children. Roughly speaking, Π-DNNF is the same as the classical DNNF although its operators are max and min instead of ∨ and ∧, respectively.*

**Example 1.** *Figure 1 depicts a sentence in Π-DNNF. Consider the Min-node (root) in this figure. This node has two children, the first contains variables A, B while the second contains variables C, D. This node is decomposable since its two children do not share variables.*

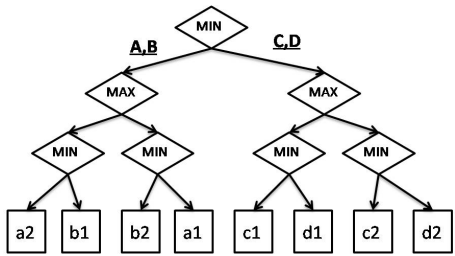

Figure 1: A sentence in Π-DNNF.

A sentence in Π-d-DNNF is a sentence in Π-DNNF satisfying decomposability and determinism (viewing ∨ and ∧ as *max* and *min* operators, respectively). A sentence in Π-sd-DNNF is a sentence in Π-DNNF satisfying decomposability, determinism and smoothness.

## 3.1 INFERENCE USING POSSIBILISTIC CIRCUITS

In (Darwiche, 2003), authors have focused on inference in compiled Bayesian networks. The main idea is based on representing the network using a *polynomial* and then retrieving answers to probabilistic queries by evaluating and differentiating the polynomial. This latter itself is exponential in size, so it has been represented efficiently using an *arithmetic circuit* that can be evaluated and differentiated in time and space linear in the circuit size. In what follows, we propose a direct adaptation of this method in the possibilistic setting. Given a min-based possibilistic network, we first encode it using a *possibilistic function* $f_{min}$ defined by two types of variables:

- *Evidence indicators*: for each variable $X_i$ in the network , we have a variable $\lambda_{x_i}$ for each instance $x_i \in D_{X_i}$.

- *Network parameters*: for each variable $X_i$ and its parents $U_i$ in the network, we have a variable $\theta_{x_i|u_i}$ for each instance $x_i \in D_{X_i}$ and $u_i \in D_{U_i}$.

$$f_{min} = \max_{\mathbf{x}} \min_{(x_i, u_i) \sim \mathbf{x}} \lambda_{x_i} \theta_{x_i|u_i} \qquad (5)$$

where $\mathbf{x}$ represents instantiations of all network variables and $u_i \sim \mathbf{x}$ denotes the compatibility relationship among $u_i$ and $\mathbf{x}$. The possibilistic function $f_{min}$ of a possibilistic network represents the possibility distribution and allows to compute possibility degrees of variables of interest. Namely, for any piece of evidence $e$ which is an instantiation of some variables $E$ in the network, we can instantiate $f_{min}$ as it returns the possibility of $e$, $\Pi(e)$ (Definition 2 and Proposition 1).

**Definition 2.** *The value of the possibilistic function $f_{min}$ at evidence $e$, denoted by $f_{min}(e)$, is the result of replacing each evidence indicator $\lambda_{x_i}$ in $f_{min}$ with 1 if $x_i$ is consistent with $e$, and with 0 otherwise.*

**Proposition 1.** *Let $\Pi G_{min}$ be a possibilistic network representing the possibility distribution $\pi$ and having the possibilistic function $f_{min}$. For any evidence $e$, we have $f_{min}(e) = \pi(e)$.*

Let figure 2 be the min-based possibilistic network used throughout the paper.

The possibilistic function of the network in figure 2 has 8 terms corresponding to the 8 instantiations of variables $F, B, D$. Two of these terms are as follows:

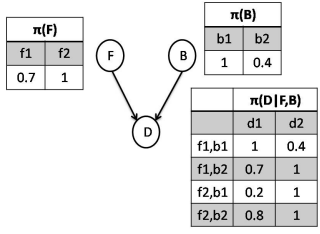

Figure 2: Example of $\Pi G_{min}$.

$f_{min} = \max(min(\lambda_{d_1}, \lambda_{f_1}, \lambda_{b_1}, \theta_{d_1|f_1,b_1}, \theta_{f_1}, \theta_{b_1}); \ min$
$(\lambda_{d_1}, \lambda_{f_2}, \lambda_{b_1}, \theta_{d_1|f_2,b_1}, \theta_{f_2}, \theta_{b_1}); \cdots )$

If the evidence $e = (d_1, b_1)$ then $f_{min}(d_1, b_1)$ is obtained by applying the following substitutions to $f_{min}$: $\lambda_{d_1} = 1, \lambda_{d_2} = 0, \lambda_{b_1} = 1, \lambda_{b_2} = 0, \lambda_{f_1} = \lambda_{f_2} = 1$. This leads to $\Pi(e) = 0.7$.

The possibilistic function $f_{min}$ is then encoded on a propositional theory (CNF) using $\lambda_{x_i}$ and $\theta_{x_i|u_i}$. For each network variable $X_i$, the encoding contains the following clauses:

$$\lambda_{x_i} \vee \lambda_{x_j} \qquad (6)$$

$$\neg\lambda_{x_i} \vee \neg\lambda_{x_j}, i \neq j \qquad (7)$$

Moreover, for each propositional variable $\theta_{x_i|u_i}$, the encoding contains the clause:

$$\lambda_{x_i} \wedge \lambda_{u_{i1}} \wedge \ldots \wedge \lambda_{u_{im}} \leftrightarrow \theta_{x_i|u_i} \qquad (8)$$

The CNF encoding, denoted by $K_{f_{min}}$ recovers the min-joint possibility distribution (proposition 2).

**Proposition 2.** *The CNF encoding $K_{f_{min}}$ of a possibilistic network encodes the joint distribution of given network.*

Once the CNF encoding is accomplished, it is then compiled into a $\Pi$-DNNF, from which we extract the possibilistic circuit $\zeta_p$ (definition 3) that implements the encoded $f_{min}$.

**Definition 3.** *A possibilistic circuit $\zeta_p$ encoded by a $\Pi$-DNNF sentence $\xi^c$ is a DAG in which leaf nodes correspond to circuit inputs, internal nodes correspond to max and min operators, and the root corresponds to the circuit output.*

As in the probabilistic case (Darwiche, 2003), this circuit can be used for linear-time inference. More precisely, computing the possibility degree of an event consists on evaluating $\zeta_p$ by setting each evidence indicator $\lambda_x$ to 1 if the event is consistent with $x$, to 0 otherwise and applying operators in a bottom-up way. This possibility degree corresponds exactly to the one computed from the min-joint possibility distribution (proposition 3). This method referred to $\Pi$-DNNF$_{PF}$ is outlined by algorithm 1. Note that the suffix $PF$ is added to signify that this method uses a possibilistic function ($f_{min}$) before ensuring the CNF encoding.

**Algorithm 1:** Inference using $\Pi$-DNNF ($\Pi$-DNNF$_{PF}$)

Data: $\Pi G_{min}$ , instance of interest $x$, evidence $e$
Result: $\Pi(x|e)$
**begin**
    **Compilation into $\Pi$-DNNF**
    Encode $\Pi G_{min}$ into $f_{min}$ using equation 5
    EncodeCNF of $\Pi G_{min}$ into $\xi$ using equations 6, 7, 8
    Compile $\xi$ into $\xi^c$
    $\zeta_p \leftarrow$ Possibilistic Circuit of $\xi^c$
    **Inference**
    Applying Operators on $\zeta_p$
    $\Pi(x, e) \leftarrow$ Root Value $(\zeta_p; (x,e))$
    $\Pi(e) \leftarrow$ Root Value $(\zeta_p; e)$
    **if** $\Pi(x, e) \prec \Pi(e)$ **then** $\Pi(x|e) \leftarrow \Pi(x, e)$
    **else** $\Pi(x|e) \leftarrow 1$
    **return** $\Pi(x|e)$
**end**

**Proposition 3.** *Let $\Pi G_{min}$ be a possibilistic network. Let $\pi_{min}$ be a joint distribution obtained by chain rule. Then for any $a \in D_a$ and $e \in D_E$, we have $\Pi(A = a|E = e) = \Pi_{min}(A = a|E = e)$ where $\Pi_{min}(A = a|E = e)$ is obtained from $\pi_{min}$ using equation 1 and $\Pi(A = a|E = e)$ is obtained from algorithm 1.*

The key point to observe here is that this approach can handle possibilistic circuits of manageable size as in the probabilistic case since some possibility values may have some specific values; for instance, whether they are equal to 0 or 1, and whether some possibilities are equal. In this case, we can say that the network exhibit some *local structure*. By exploiting it, the produced circuits can be smaller. In fact, the normalization constraint relative to the initial network will mean that we will have several values equal to 1. Thus the idea is to make an advantage from such a local structure which has a particular behavior with the max operator in order to construct more compact possibilistic circuits w.r.t. standard ones as stated by the following proposition:

**Proposition 4.** *Let $Nb_{poss}$ and $Nb_{proba}$ be the number of clauses in the possibilistic and probabilistic cases, respectively. Then $Nb_{poss} \leq Nb_{proba}$.*

Note that for particular situations where probability values are 1 or 0, we have $Nb_{poss} = Nb_{proba}$, otherwise $Nb_{poss} \prec Nb_{proba}$.

**Example 2.** *To illustrate algorithm 1 we will consider the min-based possibilistic network represented in figure 2. We are looking for $\Pi(f_2|d_1)$ with $f_2$ as instance of interest and $d_1$ as evidence. First, we encode the network as a possibilistic function and encode it on CNF. This latter is then compiled into $\Pi$-DNNF from which a possibilistic circuit is extracted. The possibility degree $\Pi(f_2|d_1)$ is computed using this circuit in polynomial time. For instance, $\Pi(f_2, d_1)$ is computed using $\zeta_p$ by just replacing*

$\lambda_{f_2} = \lambda_{d_1} = \lambda_{b_1} = \lambda_{b_2} = 1$ and applying possibilistic operators in a bottom-up way as shown in figure 3. Hence, $\Pi(f_2|d_1) = \Pi(f_2, d_1) = 0.4$ since $\Pi(f_2, d_1) = 0.4 \prec 1$.

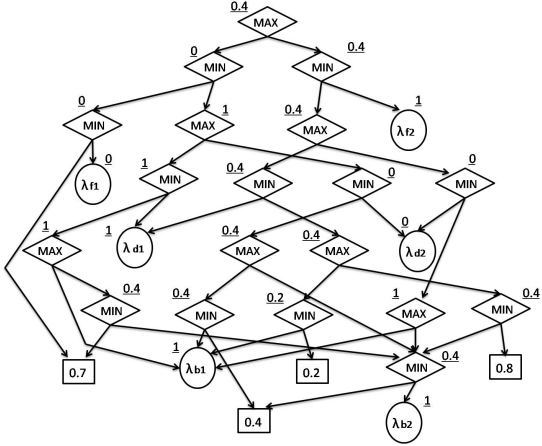

Figure 3: Inference using the possibilistic circuit ($\zeta_p$).

## 3.2 INFERENCE USING POSSIBILISTIC COMPILED REPRESENTATIONS

DNNF plays an interesting role in compiling propositional knowledge bases. It has been used to compile probabilistic networks. More precisely in (Wachter and Haenni, 2007), authors have been interested in performing a CNF logical encoding of the probability distribution induced by a bayesian network, then a compilation phase from CNF to d-DNNF. In this section, we propose to adapt this encoding in the possibilistic setting by taking into consideration the local structure aspect. This allows to reduce the number of additional variables comparing to the probabilistic encoding. Let $\Delta$ be propositions linked to network's variables and let $\theta$ be propositions linked to possibility distribution entries (equal to 1). We start by looking at the possibility distribution encoding. The logical representation of a network variable $X_i$ is defined by: $\psi_{X_i} =$

$$\bigwedge_{u_i}\left(\bigwedge_{\theta_{x_i|u_i} \in \Omega_{\theta_{X_i|u_i}}} \left(u_{i1} \wedge \cdots \wedge u_{im} \wedge \theta_{x_i|u_i} \rightarrow x_i\right)\right) \quad (9)$$

By taking the conjunction of all logical representations of variables, we obtain the network's representation $\psi$ as follows:

$$\psi = \bigwedge_{X_i \in \Delta} \psi_{X_i} \quad (10)$$

The CNF encoding, denoted by $K_\psi$ indeed recovers the min-joint possibility distribution (proposition 5).

**Proposition 5.** *Let $\pi_{min}$ be the joint possibility distribution obtained using the chain rule with the minimum operator and $\Pi$ be the possibility degree computed* from a function $f_\psi$ encoding the CNF. Then, we have $\pi_{min}(x_i, ..., x_j) = \Pi(x_i, ..., x_j)$, i.e. $f_\psi$ recovers the min-joint possibility distribution $\pi_{min}$.

Comparing theoretically the probabilistic and the possibilistic case allows us to deduce the following proposition:

**Proposition 6.** *The possibilistic encoding of a possibilistic network given by $K_\psi$ (equation 10) is more compact than the probabilistic encoding given in (Wachter and Haenni, 2007).*
*In fact, the number of variables used in $K_\psi$ is less than the one used in (Wachter and Haenni, 2007). In particular for parameters, our approach uses one variable per different weight, while in the probabilistic encoding one variable per parameter. For each clause in $K_\psi$ there exists a clause of the same size in the probabilistic encoding. The converse is false.*

Once the qualitative network is encoded by $K_\psi$, it is compiled into a compilation language that supports the transformations *conditioning* and *forgetting* and the query *possibilistic computation*. This language is $\Pi$-DNNF (proposition 7). Therefore, the CNF encoding is first compiled, and the resulting $\Pi$-DNNF is then used to compute efficiently, i.e. in polynomial time a-posteriori possibility degrees (proposition 8). This method referred to $\Pi$-DNNF is outlined by algo. 2.

**Proposition 7.** *$\Pi$-DNNF supports conditioning, forgetting and possibilistic computation.*

---

**Algorithm 2:** Inference using $\Pi$-DNNF

Data: $\Pi G_{min}$ , instance of interest $x$, evidence $e$
Result: $\Pi(x|e)$
**begin**
    **Compilation into $\Pi$-DNNF**
    EncodeCNF of $\Pi G_{min}$ into $\psi$ using equation 10
    Compile $\psi$ into $\psi_p^c$
    **Inference**
    $v_1 \leftarrow$ Explore $\Pi$-DNNF(x $\wedge$ e, $\psi_p^c$)
    $v_2 \leftarrow$ Explore $\Pi$-DNNF(e, $\psi_p^c$)
    **if** $v_1 \prec v_2$ **then** $\Pi(x|e) \leftarrow v_1$ **else** $\Pi(x|e) \leftarrow 1$
    **return** $\Pi(x|e)$
**end**

---

**Proposition 8.** *Let $\Pi G_{min}$ be a possibilistic network. Let $\pi_{min}$ be a joint distribution obtained by chain rule. Then for any $a \in D_a$ and $e \in D_E$, we have $\Pi(A = a|E = e) = \Pi_{min}(A = a|E = e)$ where $\Pi_{min}(A = a|E = e)$ is obtained from $\pi_{min}$ using equation 1 and $\Pi(A = a|E = e)$ is obtained from algorithm 2.*

**Example 3.** *Let us illustrate algorithm 2. In fact, $\psi$ of the network of figure 2 is : $\psi = \psi_F \wedge \psi_B \wedge \psi_D = \{(\theta_1 \vee f_2) \wedge (\theta_2 \vee b_1) \wedge (f_2 \vee b_2 \vee \theta_2 \vee d_1) \wedge (f_2 \vee b_1 \vee \theta_1 \vee d_2) \wedge (f_1 \vee b_2 \vee \theta_3 \vee d_2) \wedge (f_1 \vee b_1 \vee \theta_4 \vee d_2)\}$ such as $\theta_1$, $\theta_2$, $\theta_3$ and $\theta_4$ correspond respectively to 0.8, 0.7, 0.4 and 0.2.*

*To compute $\Pi(f_2|d_1)$, we should first compute $\Pi(f_2, d_1)$ using algorithm 3. The first step is to check if we have at least*

**Algorithm 3:** Explore Π-DNNF

Data: a set of instances $x$, compiled representation $\psi_p^c$

Result: $\Pi(x)$

**begin**
    **if** $\forall\ x_i \in x,\ \theta_{x_i|U_i}$ *is not a leaf node* **then**
        $\Pi(x) \leftarrow 1$
    **else**
        y= $\{x_i \mid \forall,\ \theta_{x_i|U_i}$ is a leaf node $\forall\ U_i \subseteq$ x$\}$
        $\psi_{p|y}^c \leftarrow$ Condition $\psi_p^c$ on $y$
        $\psi_{p\downarrow|y}^c \leftarrow$ Forget $\Delta$ from $\psi_{p|y}^c$
        Applying Operators on $\psi_{p\downarrow|y}^c$
        $\Pi(x) \leftarrow$ Root Value of $\psi_{p\downarrow|y}^c$
    **return** $\Pi(x)$
**end**

one $\theta$ as a leaf node. In this example, we have $\theta_{d_1|f_2,b_1}$ and $\theta_{d_1|f_2,b_2}$ as leaf nodes, hence conditioning should be performed. Then, a computation step is required by applying in a bottom-up way Min and Max operators on the forgotten Π-DNNF. Therefore, $\Pi(f_2|d_1) = \Pi(f_2, d_1) = 0.4$.

## 4 NEW POSSIBILISTIC INFERENCE ALGORITHM

In (Benferhat et al., 2002), authors have been interested in the transition of possibilistic networks into possibilistic logic bases. The starting point is that the possibilistic base associated to a possibilistic network is the result of the fusion of elementary bases. Definition 4 presents the transformation of a min-based possibilistic network into a possibilistic knowledge base.

**Definition 4.** *A binary variable $X_i$ of a possibilistic network can be expressed by a local possibilistic knowledge base as follows:* $\Sigma_{X_i} = \{(\neg x_i \vee \neg u_i, \alpha_i) : \alpha_i = 1 - \pi(x_i|u_i) \neq 0\}$. *The possibilistic knowledge base of the whole network is:* $\Sigma_{min} = \Sigma_{X_1} \cup \Sigma_{X_2} \cup \cdots \cup \Sigma_{X_n}$.

In another angle, researchers in (Benferhat et al., 2007) have focused on the compilation of bases under the possibilistic logic policy in order to be able to process inference from it in a polynomial time. The combination of these methods allows us to propose a new alternative approach to possibilistic inference. This is justified by the fact that the possibilistic logic reasoning machinery can be applied to directed possibilistic networks (Benferhat et al., 2002).

The idea is to encode the possibilistic knowledge base $\Sigma_{min}$ into a classical propositional base (CNF). Let $A = \{a_1, ..., a_n\}$ with $a_1 \succ ... \succ a_n$ the different weights used in $\Sigma_{min}$. A set of additional propositional variables, denoted by $A_i$, which correspond exactly to the number of different weights, are incorporated and for each formula $\phi_i, a_i$ will correspond the propositional formula $\phi_i \vee A_i$. Hence, the propositional

encoding of $\Sigma_{min}$, denoted by $K_{\Sigma}$ is defined by:

$$K_{\Sigma} = \{\phi_i \vee A_i : (\phi_i, a_i) \in \Sigma_{min}\} \qquad (11)$$

The following proposition shows that the CNF encoding $K_{\Sigma}$ recovers the min-joint possibility distribution.

**Proposition 9.** *Let $\pi_{min}$ be the joint possibility distribution obtained using the chain rule with the minimum-based conditioning and let $K_{\Sigma}$ be the propositional base associated with the possibilistic network given by equation 11. Let $\phi_i$ be a propositional formula associated with a degree $a_i$. Then $\forall \omega \in \Omega$, $\Pi(\omega) = 1$ iff $\{\neg A_1, ..., \neg A_n\} \wedge \omega \wedge K_{\Sigma}$ is consistent. $\Pi(\omega) = a_i$ iff $\{\neg A_1, ..., \neg A_i\} \wedge \omega \wedge K_{\Sigma}$ is inconsistent and $\{\neg A_1, ..., \neg A_{i-1}\} \wedge \omega \wedge K_{\Sigma}$ is consistent.*

The CNF encoding $K_{\Sigma}$ is then compiled into a target compilation language in order to compute a-posteriori possibility degrees in an efficient way. Here, we are interested in a particular query useful for possibilistic networks, namely *what is the possibility degree of an event $A = a$ given an evidence $E = e$?* Therefore, we propose to adapt the algorithm given in (Benferhat et al., 2007) in order to respond to this query as shown by algorithm 4. Proposition 10 shows that the possibility degree computed using algorithm 4 and the one computed using the min-based joint possibility distribution are equal. Note that this approach is qualified to be flexible since it takes advantage of existing propositional knowledge bases compilation methods (Benferhat et al., 2007). This method referred to DNNF-PKB is outlined by algorithm 4.

**Algorithm 4:** Inference using DNNF

Data: $\Pi G_{min}$ , instance of interest $x$, evidence $e$

Result: $\Pi(x|e)$

**begin**
    **Transformation into $K_{\Sigma}$**
    Transform $\Pi G_{min}$ into $\Sigma_{min}$ using definition 4
    Transform $\Sigma_{min}$ into $K_{\Sigma}$ using equation 11
    **Inference**
    $K_{\Sigma}^c \leftarrow Target(K_{\Sigma})$
    $K \leftarrow K_{\Sigma}^c$
    StopCompute $\leftarrow$ false
    i $\leftarrow 1$
    $\Pi(x|e) \leftarrow 1$
    **while** *(K $\nvdash$ $A_i \vee \neg e$) and (i $\leq$ k) and (StopCompute=false)* **do**
        K $\leftarrow$ condition (K, $\neg A_i$)
        **if** $K \vDash \neg x$ **then**
            $StopCompute \leftarrow$ true
            $\Pi(x|e) \leftarrow$ 1-$degree(i)$
        **else**
            $i \leftarrow i+1$
    **return** $\Pi(x|e)$
**end**

**Proposition 10.** *Let $\Pi G_{min}$ be a possibilistic network. Let $\pi_{min}$ be a joint distribution obtained by*

*chain rule. Then for any $a \in D_a$ and $e \in D_E$, we have $\Pi(A = a | E = e) = \Pi_{min}(A = a | E = e)$ where $\Pi_{min}(A = a | E = e)$ is obtained from $\pi_{min}$ using equation 1 and $\Pi(A = a | E = e)$ is obtained from Algo. 4.*

**Example 4.** *To illustrate algorithm 4 we will consider the min-based possibilistic network represented in figure 2.*

*The CNF encoding is as follows : $K_\Sigma = (d_2 \vee f_1 \vee b_2 \vee A_1), (b_1 \vee A_2), (d_1 \vee f_2 \vee b_2 \vee A_2), (f_2 \vee A_3), (d_2 \vee f_2 \vee b_1 \vee A_3), (d_2 \vee f_1 \vee b_1 \vee A_4)$ such as $A_1$ (0.8), $A_2$ (0.6), $A_3$(0.3) and $A_4$ (0.2) are propositional variables followed by their weights under brackets. Compiling $K_\Sigma$ into DNNF results in: $K_\Sigma^C = ((b_2 \wedge A_2) \wedge [(A_3 \wedge f_1) \vee (f_2 \wedge [d_2 \vee (A_4 \wedge d_1)])]) \vee (b_1 \wedge [[f_2 \wedge (d_2 \vee (A_1 \vee d_1))] \vee [(f1 \wedge A_3) \wedge (d_1 \vee (A_2 \wedge d_2))]]))$. The computation of $\Pi(f_2 | d_1)$ using $K_\Sigma^C$ requires two iterations. Therefore, $\Pi(f_2 | d_1) = 1 - degree(2) = 0.4$.*

Due to the compilation step, this algorithm runs in polynomial time. Moreover, the number of additional variables is low since it corresponds exactly to the number of priority levels existing in the base.

# 5 COMPARATIVE AND EXPERIMENTAL STUDIES

The paper analyzes three compilation-based methods, namely DNNF-PKB, $\Pi$-DNNF and $\Pi$-DNNF$_{PF}$. The first dimension that differentiates the three approaches proposed in this paper is the CNF encoding. It consists of specifying the number of variables and clauses per approach.

The CNF of DNNF-PKB is based on encoding $\neg x$ where $x$ is an instance of interest having a possibility degree different from 1. In $\Pi$-DNNF, we write implications relative to instances having 1 as possibility degree. We can notice that the local structure in both methods is exploited in semantically different ways. In DNNF-PKB, the encoding uses the number of different weights as the number of additional variables while the $\Pi$-DNNF encoding uses the number of the non-redundant possibility degrees different from 1 in the distributions. Regarding the number of clauses, both methods handle possibility degrees different from 1. This leads us to the following proposition:

**Proposition 11.** *The CNF encodings of DNNF-PKB and $\Pi$-DNNF have the same number of variables and clauses.*

The CNF encoding of $\Pi$-DNNF$_{PF}$ is different from the ones of DNNF-PKB and $\Pi$-DNNF. Proposition 12 shows the difference between $\Pi$-DNNF$_{PF}$ and DNNF-PKB in terms of number of variables and clauses.

**Proposition 12.** *The number of variables and clauses in $\Pi$-DNNF$_{PF}$ is more important than those in DNNF-PKB.*

Indeed, in $\Pi$-DNNF$_{PF}$, we associate propositional variables not only to possibility degrees (parameters), but also to each value $x_i$ of $X_i$. While in DNNF-PKB only $m$ new variables are added (one variable per different degree).

Let us now analyze these three approaches from experimental points of view. Our experimentation is performed on random possibilistic networks. More precisely, we have compared DNNF-PKB and $\Pi$-DNNF$_{PF}$ on 100 possibilistic networks having from 10 to 50 nodes. As mentioned that the approaches focus mainly on encoding the possibilistic network as a CNF then compile it into the appropriate language, hence, it should be interesting to compare the CNF parameters (the number of variables and clauses) and the DNNF parameters (the number of nodes and edges) for the two methods.

## 5.1 CNF PARAMETERS

First we propose to test the CNF encodings characterized by the number of variables and the number of clauses. Regarding DNNF-PKB, the number of additional variables correspond to the number of weights which are different. While in $\Pi$-DNNF$_{PF}$, variables are both those associated to the possibility degrees of each distribution and those to variable's instances. The number of clauses for each method is related to the CNF encoding itself. Figure 4 shows the results of this experimentation. Each approach is characterized by a curve for the average number of variables and a curve for the average number of clauses. It is clear that the higher the number of nodes considered in the possibilistic network, the higher the number of variables and clauses. Figure 4 shows that DNNF-PKB has the lower number of variables and clauses comparing to $\Pi$-DNNF$_{PF}$, which confirms the theoretical results detailed above.

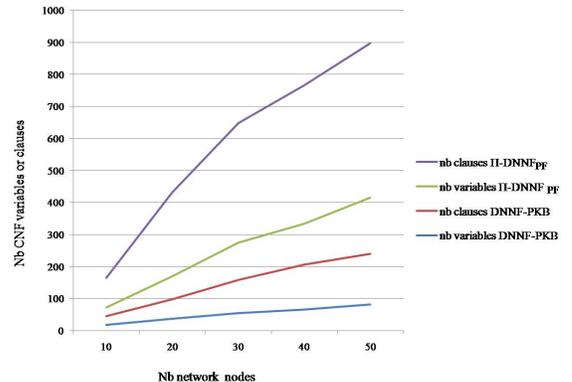

Figure 4: CNF parameters.

## 5.2 DNNF PARAMETERS

Once we obtain the CNF encodings, it is important to compare the number of nodes and edges for each compiled base. Figure 5 represents the average size of the compiled bases for the two methods in terms of nodes and edges numbers. We remark that the number of nodes and edges depends deeply on CNF parameters. More precisely, the number of nodes and edges in DNNF-PKB is considered narrow comparing to $\Pi$-DNNF$_{PF}$. This can be explained by the lower number of variables and clauses on CNFs and the local structure which shrinks the sizes of compiled bases. Comparing DNNF-PKB to $\Pi$-DNNF$_{PF}$, the behavior of DNNF-PKB is important.

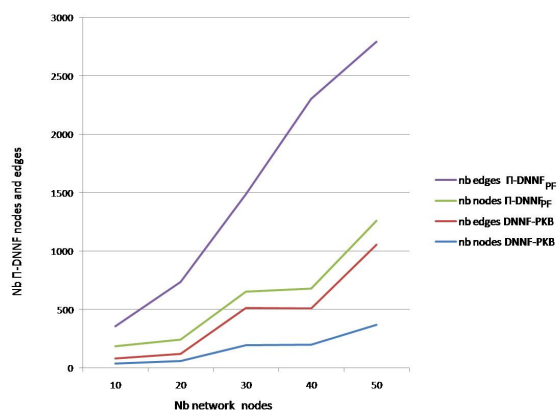

Figure 5: DNNF parameters.

## 6 CONCLUSION

This paper proposes algorithms that ensure inference in possibilistic networks using compilation techniques. First, we have proposed possibilistic adaptations of two compilation-based probabilistic methods, namely $\Pi$-DNNF$_{PF}$ and $\Pi$-DNNF. Then we have developed a new possibilistic inference method DNNF-PKB based on a transformation phase from a possibilistic network into a compiled possibilistic knowledge base. We theoretically show that DNNF-PKB and $\Pi$-DNNF share the same number of variables and clauses even they are based on different computations in their inference process since the first is based on necessity degrees and the second on possibility degrees. We have also shown that DNNF-PKB is more compact than $\Pi$-DNNF$_{PF}$ which proves the importance of the possibilistic setting versus the probabilistic setting. All these results were confirmed by experimental results. A future work will be to compare these algorithms with the well-known junction tree propagation algorithm. Another future work is to exploit results of this paper in order to infer efficiently interventions in possibilistic causal networks (Pearl, 2000) (Benferhat and Smaoui, 2007).

**Acknowledgements**

We thank the anonymous reviewers for many interesting comments and suggestions. Also, we wish to thank Mark Chavira for our valuable discussions on this subject. The third author would like to thank the project ANR Placid.